\documentclass[runningheads]{llncs}
\usepackage[margin=3 cm]{geometry}
\usepackage[T1]{fontenc}
\usepackage{graphicx}
\usepackage{booktabs}
\usepackage{multirow}
\usepackage{amsmath}
\usepackage{amssymb}
\usepackage[authoryear]{natbib}

\usepackage{placeins}

\begin{document}

\title{Beyond the Black Box: Interpretable Models of Human Randomisation Failures}
\titlerunning{Beyond the Black Box}
\author{Ngoc Linh Dao}
\authorrunning{Ngoc Linh Dao}
\institute{Alpen-Adria-University of Klagenfurt, Austria \\ }
\maketitle

\begin{abstract}
Mixed-strategy equilibrium predicts i.i.d play: past actions should not help predict future decisions. Human players, however, systematically depart from this benchmark, and in O'Neill's zero-sum card game, these departures can be predicted by black-box sequence models such as LSTMs. This paper asks whether that predictive power can be achieved by transparent alternatives that also reveal the behavioural structure behind it. Using 84,060 decisions from 2,802 pairs, the analysis first benchmarks naive and behavioural models against interpretable machine-learning and deep-learning models, then evaluates the modified EWA specifications of prior work against these benchmarks and uses the LASSO diagnostics to motivate a further nested frequency-tracking extension. The results show that repeat/avoid behaviour, especially players' management of their own recent action histories, accounts for most of the interpretable and strategically exploitable signal, while frequency tracking adds little out of sample.
\end{abstract}

\section{Introduction}\label{sec:intro}

Being unpredictable is a strategic skill, and a difficult one. When asked to
randomise, people often over-alternate, avoid recent choices, and leave
history-dependent traces. O'Neill's four-card game makes this failure
measurable. In each round, a Red player and a Black player simultaneously choose
one of four cards, $C=\{1,2,3,K\}$. Red wins if both players choose $K$, or if
both choose different number cards. Black wins if exactly one player chooses
$K$, or if both players choose the same number card. The game is zero-sum and
has a sharp mixed-strategy benchmark: in equilibrium, each player chooses each
number card with probability $0.2$ and the face card $K$ with probability
$0.4$, independently across rounds. Thus, equilibrium play is serially
independent: any history-dependent predictability is therefore a departure from the equilibrium benchmark. Because the equilibrium is asymmetric, with Red winning
with probability $0.4$ and Black with probability $0.6$, all models are
estimated separately by role.

\paragraph{}Prior evidence shows a consistent tension. \citet{oneill1987} found that
aggregate play in O'Neill's game was close to the minimax benchmark. However,
\citet{brown1990} showed that individual choices remain history-dependent:
players' own past actions and their opponents' recent histories help predict
future choices. Such structure has motivated behavioural learning models,
including reinforcement learning~\citep{rotherev1995}, fictitious
play~\citep{brown1949}, and Experience-Weighted Attraction
(EWA)~\citep{camerer1999}. More recently, machine-learning approaches have
brought an out-of-sample prediction perspective to economic
behaviour~\citep{mullainathan2017,athey2019}. Most directly,
\citet{hirasawa2025} show in repeated O'Neill play that flexible sequence
models predict human deviations well.

\paragraph{}The remaining question is what this predictability is made of: Is it
own-action habits, opponent tracking, payoff learning, or frequency tracking?
Following \cite{rudin2019}, "the way forward is to design models that are inherently interpretable", we use 84{,}060 decisions from 2{,}802 pairs over 30 rounds, to construct both neural benchmarks and interpretable alternatives in a one-step-ahead prediction task: given history through round $t-1$, predict a player's card at round $t$. We (i) replicate the interpretable-versus-black-box comparison; (ii) enrich the feature analysis with additional history variables; and 
(iii) extend modified EWA with ME3, an ML-guided frequency-tracking term motivated by the frequency features the LASSO selects, testing whether players best-respond to the opponent's recent modal card.

\begin{figure}
    \centering
    \includegraphics[width=0.4\linewidth]{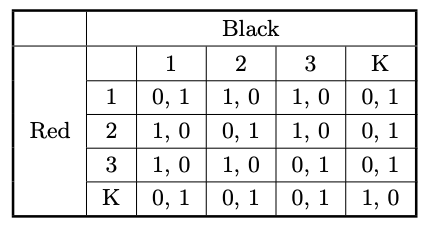}
    \caption{Payoff Matrix of the O'Neill's Card game}
    \label{fig:placeholder}
\end{figure}
\section{Methodology}\label{sec:method}

\subsection{General Formulation}
All models share the same prediction task: for each player role $i$, predict the
next card $a_i^t$ from the history $h_t$ available before round $t$. Each produces probabilities over the four cards through a multinomial logit $P_i^{a}(t)\propto\exp\{\beta_{i,a}^{\top}x_i(h_t)\}$, and the models differ only in how they summarise $h_t$: as constants,
lagged actions, payoff-based attractions, sparse history features, or learned
sequence representations.

\subsection{Econometric baselines}
The empirical-constant (i.i.d.) model uses only constants, so its predicted
probabilities are the empirical card frequencies and do not vary with history,
$P_i^{a}(t)\propto\exp\{c_i^{a}\}$. To allow basic history dependence, we also
estimate serial-correlation models in which the probability of card $a$ shifts
if the player chose $a$ in one of the previous $n$ rounds. We also employ the restricted version (number-card coefficients shared) and report
orders $1$ and $4$; among tested lengths, order $4$ predicts best.

\subsection{Experience-Weighted Attraction (EWA)}
EWA~\cite{camerer1999} models learning by assigning each card an attraction
and updating it after each round. With experience weight
$N_i(t)=\rho_i N_i(t-1)+1$,
\begin{equation}
A_i^{a}(t)=
\frac{
\phi_i N_i(t{-}1)A_i^{a}(t{-}1)
+
\left[\delta_i+(1-\delta_i)\mathbf{1}\{a_i^{t}=a\}\right]
\pi_i(a,a_{-i}^{t})
}{
N_i(t)
}.
\end{equation}
where $\rho_i$ discounts past experience, $\phi_i$ past attractions, and
$\delta_i$ weights foregone payoffs. Choices follow a softmax,
$P_i^{a}(t{+}1)\propto\exp\{\lambda_i A_i^{a}(t)\}$, with sensitivity
$\lambda_i$; initial attractions $A_i^{a}(0)$ and weight $N_i(0)$ capture
predispositions. Parameters are estimated by maximum likelihood (L-BFGS-B,
imposing $\rho_i,\delta_i\in[0,1]$, $\lambda_i\ge0$); the model nests
reinforcement and belief learning as special cases.

\subsection{Machine Learning models}
We estimate two interpretable machine-learning models and two neural
benchmarks. The LASSO multinomial logit and decision tree use the same enriched
history-based feature set, including lagged actions, streaks, joint action
profiles, win-loss histories, recent card counts, and best-response indicators.
The DNN and LSTM replicate Hirasawa et al.~\cite{hirasawa2025}: the DNN uses a
fixed 4-recent-history encoding, while the LSTM processes the action sequence
recurrently.  Both output a softmax over the four cards, trained with
Adam and hyperparameters selected within the training data. They serve not as
behavioural explanations but as high-capacity benchmarks for how much history-
dependent structure is extractable

\subsection{Modified EWA family}
\begin{figure}
    \centering
    \includegraphics[width=0.5\linewidth]{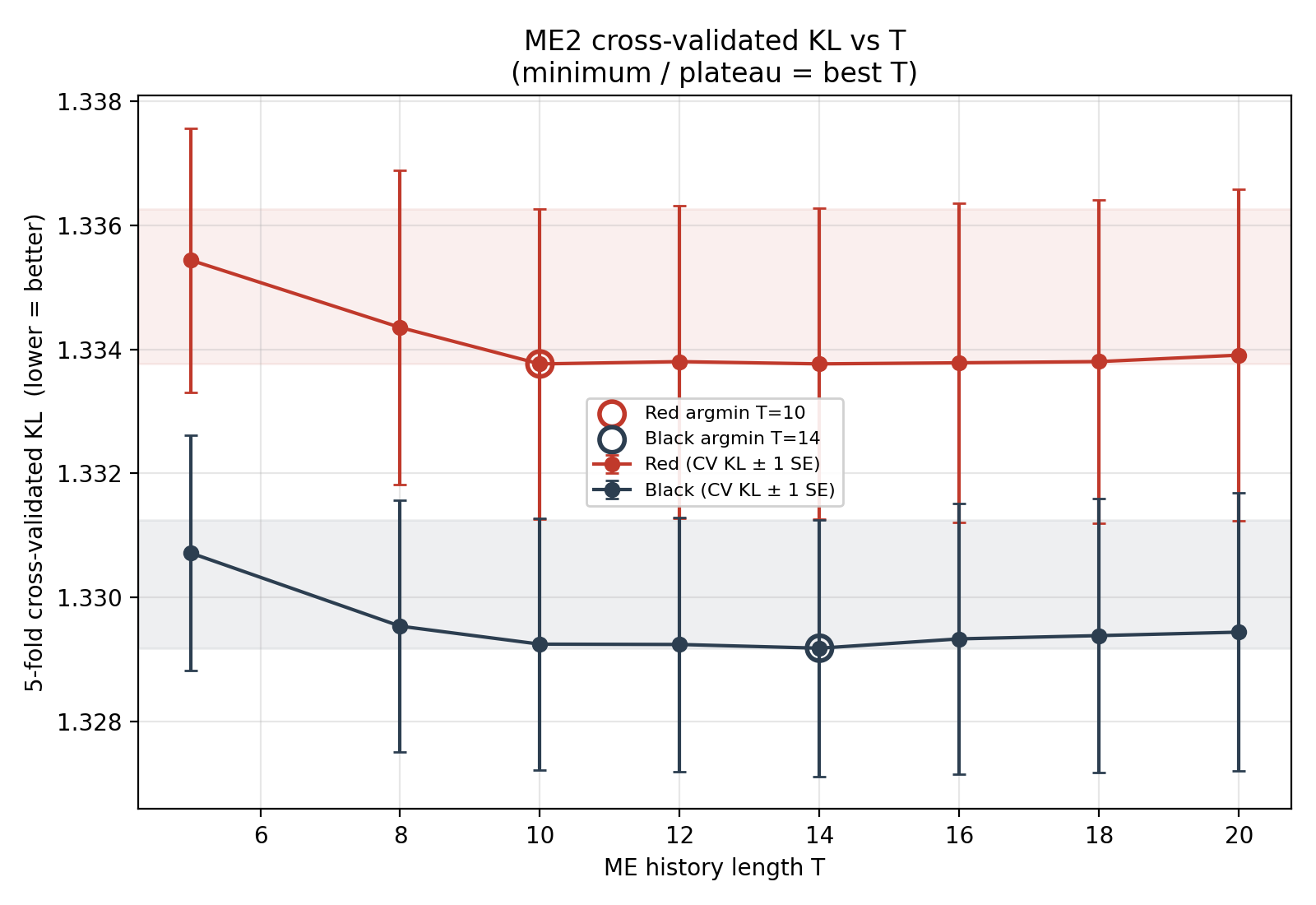}
    \caption{ME2's KL best T trajectory}
    \label{fig:tsweep}
\end{figure}

The modified EWA family combines mechanisms of two provenances. ME1 and ME2 follow~\cite{hirasawa2025}, who specified own- and opponent-repeat/avoid terms from behavioural theory, prior to and independently of the feature analysis here. ME3 is our extension: it translates the frequency-tracking features selected by the LASSO into a parsimonious behavioural term, and is therefore ML-guided by construction. All modified EWA models share the choice rule
$P_i^{a}(t)\propto\exp\{\lambda_i A_i^{a}(t-1)+M_i^{a}(t)\}$, where $A_i^{a}$
is the EWA attraction and $M_i^{a}(t)$ collects additional history terms. The family nests: ME1 adds own
repeat/avoid terms; ME2~\cite{hirasawa2025} adds an opponent term over
$W_i^{a}$, the opponent cards that $a$ beats; and ME3 (our extension) adds a
frequency-tracking term:
\begin{equation}
M_i^{a}(t)=\underbrace{\textstyle\sum_{\tau=1}^{T}\alpha_i^{a\tau}R_i^{t}(\{a\},\tau)}_{\text{ME1: own repeat/avoid}}
+\underbrace{\textstyle\sum_{\tau=1}^{T}\gamma_i^{a\tau}R_{-i}^{t}(W_i^{a},\tau)}_{\text{ME2: opponent repeat/avoid}}
+\underbrace{\textstyle\sum_{w\in\{3,4,5\}}\beta_i^{aw}F_i^{t}(a,w)}_{\text{ME3: frequency-tracking}},
\label{eq:me3}
\end{equation}
where the frequency indicator
$F_i^{t}(a,w)=\mathbf{1}\{a\in\mathrm{BR}_i(\mathrm{mode}(a_{-i}^{t-1},\dots,a_{-i}^{t-w}))\}$
equals one if $a$ best-responds to the opponent's modal card over the last
$w$ rounds. We use short windows $w\in\{3,4,5\}$ for modal tracking assuming identifying the opponent's dominant card requires less temporal accumulation than learning repeat/avoid patterns; cross-validation of $w$ is reserved as future work, but preliminary exploration suggested no substantial sensitivity to this range. The starred variants ($\text{ME2}^{*}$,
$\text{ME3}^{*}$) extend the memory length $T$, selected by cross-validation:
out-of-sample KL forms a flat plateau (Fig.~\ref{fig:tsweep}) with optima at
$T{=}10$ (red) and $T{=}14$ (black), so $T$ is only weakly identified. All
models are fit by maximum likelihood; the high-dimensional and extended-memory
variants use a JAX-based optimiser with multiple random starts, which makes
the otherwise prohibitive fits tractable and well-converged.
\section{Results and Discussion}\label{sec:results}

\subsection{Evaluation Metrics.}
We evaluate predictions using three metrics:
\begin{itemize}
    \item \textbf{KL divergence} For a predicted distribution $P$ over the four
    cards and realised action $a'$, the loss is $-\log P(a')$. Summed over
    observations, this is the negative log-likelihood, so the KL metric matches
    the maximum-likelihood objective used to fit the models.

    \item \textbf{Strategic error rate (SER).} SER can be understood as the win rate of the focal player if an opponent best-responded to the model's predicted
    strategy. Lower SER means the prediction is more strategically exploitable.

    \item \textbf{Relative completeness (RC).} RC measures the fraction of the
    LSTM's KL improvement over the empirical-constant baseline recovered by a
    model:
    \[
    RC_m =
    \frac{KL_m-KL_{\mathrm{const}}}
         {KL_{\mathrm{LSTM}}-KL_{\mathrm{const}}}.
    \]
    Thus, $RC=0$ corresponds to the constant baseline and $RC=1$ to the LSTM
    benchmark.
\end{itemize}

All out-of-sample results use five-fold cross-validation partitioned by pair.

\subsection{Result interpretation}
\begin{table}[t]
\centering
\caption{Model performance by player role.}
\label{tab:main-results}
\scriptsize
\setlength{\tabcolsep}{3pt}
\resizebox{\textwidth}{!}{%
\begin{tabular}{llrrrrrrrr}
\toprule
 & & \multicolumn{4}{c}{Red player} & \multicolumn{4}{c}{Black player} \\
\cmidrule(lr){3-6}\cmidrule(lr){7-10}
Evaluation & Model & \#Param & KL & SER & RC & \#Param & KL & SER & RC \\
\midrule
\multicolumn{10}{l}{\textit{Classics}} \\
In-sample & Nash              & 0  & 1.3670 & 0.413 & -0.241 & 0  & 1.3589 & 0.590 & -0.198 \\
In-sample & Constant          & 3  & 1.3599 & 0.350 &  0.000 & 3  & 1.3539 & 0.562 &  0.000 \\

\addlinespace
\multicolumn{10}{l}{\textit{Serial correlation}} \\
In-sample & Unrestricted $t=1$ & 7  & 1.3536 & 0.350 & 0.214 & 7  & 1.3484 & 0.540 & 0.218 \\
In-sample & Unrestricted $t=4$ & 19 & 1.3508 & 0.343 & 0.310 & 19 & 1.3445 & 0.532 & 0.373 \\
In-sample & Restricted $t=1$   & 5  & 1.3536 & 0.350 & 0.214 & 5  & 1.3484 & 0.540 & 0.218 \\
In-sample & Restricted $t=4$   & 11 & 1.3509 & 0.344 & 0.306 & 11 & 1.3446 & 0.532 & 0.369 \\

\addlinespace
\multicolumn{10}{l}{\textit{Behavioural model}} \\
In-sample & EWA               & 8  & 1.3528 & 0.331 & 0.241 & 8  & 1.3484 & 0.559 & 0.218 \\

\addlinespace
\multicolumn{10}{l}{\textit{Machine-learning models}} \\
CV & LASSO                   & 537.2   & 1.3373 & 0.318 & 0.769 & 550.6   & 1.3320 & 0.521 & 0.869 \\
CV & Decision tree           & 9.2     & 1.3514 & 0.339 & 0.289 & 8.0     & 1.3449 & 0.540 & 0.357 \\
CV & DNN                     & 5{,}358 & 1.3469 & 0.330 & 0.442 & 5{,}058 & 1.3427 & 0.531 & 0.444 \\
CV & LSTM                    & 6{,}272 & 1.3305 & 0.305 & 1.000 & 5{,}610 & 1.3287 & 0.509 & 1.000 \\

\addlinespace
\multicolumn{10}{l}{\textit{Improved EWA models}} \\
CV & EWA ME1          & 24  & 1.3373 & 0.317 & 0.769 & 24  & 1.3337 & 0.519 & 0.802 \\
CV        & EWA ME2$^\ast$   & 86  & 1.3338 & 0.312 & 0.888 & 118 & 1.3292 & 0.518 & 0.980 \\
CV        & EWA ME3$^\ast$   & 98  & 1.3336 & 0.311 & 0.895 & 130 & 1.3292 & 0.518 & 0.980 \\
\bottomrule
\end{tabular}%
}
\begin{minipage}{0.95\textwidth}
\footnotesize
\textit{Notes:} For CV models, \#Param is averaged across folds where applicable.
\end{minipage}
\end{table}

Table 1 points to a clear interpretation: human play is predictable, but the predictive structure is specific. Nash performs worse than the empirical-constant model for both roles, while standard EWA recovers little of the LSTM gap for either role. Simple serial-correlation models improve on this, especially at order t=4, but still explain only a limited share of the LSTM gap. Thus, the main regularity is not simply equilibrium deviation, payoff learning, or generic short-run dependence.

\begin{table}[t]
\centering
\caption{LASSO-selected feature families and behavioural interpretation.}
\label{tab:lasso-selected-summary}
\scriptsize
\setlength{\tabcolsep}{3.5pt}
\begin{tabular}{p{0.18\linewidth}p{0.30\linewidth}p{0.30\linewidth}p{0.17\linewidth}}
\toprule
Mechanism & Red-player evidence & Black-player evidence & Interpretation \\
\midrule

Own repeat/avoid (ME1)
&
Own ``did not play card $a$'' variables are commonly selected; just-played
variables are negative for the corresponding card; long number and $K$ streaks
enter differently.
&
The same pattern appears: own avoided-card variables are positive for number
cards, just-played indicators are negative, and long number/$K$ streaks enter
with different signs.
&
Strongest channel: players manage their own action sequences. \\

\addlinespace
Opponent payoff-history tracking (ME2)
&
Variables for whether the opponent repeatedly played or avoided cards beaten by
the focal card are selected for several cards.
&
Analogous opponent repeated/avoided payoff-relevant-card variables are selected
for several cards.
&
Secondary channel: players react to payoff-relevant opponent patterns. \\

\addlinespace
Frequency tracking (ME3)
&
Best-response-to-opponent-mode indicators, opponent mode-tie indicators, and
opponent recent $K$ counts are selected, but with small coefficients.
&
Best-response-to-opponent-mode indicators, opponent mode-tie indicators, and
opponent recent $K$ counts are also selected, again with small coefficients.
&
Weak channel: modal-frequency tracking exists but is not the main driver. \\

\addlinespace
Outcome dependence
&
Recent wins/losses and card-specific outcome variables are selected.
&
Recent wins/losses and card-specific outcome variables are selected.
&
Outcomes matter, but this does not imply loss overweighting. \\

\addlinespace
Number/$K$ asymmetry
&
Long number-card streaks and long $K$ streaks enter differently.
&
Long number-card streaks and long $K$ streaks also enter differently.
&
Supports estimating number-card and $K$ behaviour separately. \\

\bottomrule
\end{tabular}
\end{table}
\paragraph{}The machine-learning benchmarks show that richer history features matter. LASSO performs strongly, reaching relative completeness of 0.769 for Red and 0.869 for Black. Since LASSO selects a sparse set of predictors, this suggests that much of the predictive signal can be recovered from interpretable history variables. By contrast, the decision tree performs only modestly, indicating that a small number of if–then rules is not enough. The LSTM remains the strongest benchmark, outperforming every traditional model by a wide margin.

\paragraph{}
The modified EWA models sharpen the behavioural interpretation. ME1, which adds
own repeat/avoid terms, produces the first major improvement over standard EWA,
reaching relative completeness of $0.769$ for Red and $0.802$ for Black. This
suggests that players' own action histories are highly predictive: subjects
appear to manage their sequences by avoiding or returning to recently used
cards. ME2$^\ast$ closes most of the remaining gap, reaching relative completeness of
$0.888$ for Red and $0.980$ for Black. This improvement should be interpreted
cautiously because ME2$^\ast$ adds payoff-relevant opponent repeat/avoid terms
and also uses longer selected memory lengths. The safer conclusion is that
extended repeat/avoid structure captures most of the predictable pattern in
play. ME3 tests whether a further opponent-history mechanism is missing: best-response to the opponent's recent modal card. The answer is mostly negative: Despite performing well with in-sample data, ME3$^\ast$ barely improves on ME2$^\ast$ for Red players and is unchanged for Black players after rounding (Table~\ref{tab:main-results}). Thus, modal-frequency tracking is plausible, but it is not the main source of predictability.
\paragraph{}
Table~\ref{tab:lasso-selected-summary} provides a model-agnostic check on this
ranking. The point is not the precise coefficient magnitudes, which depend on
scaling and regularization, but the selected feature families. The sparse
classifier selects own repeat-and-avoid variables most consistently,
payoff-relevant opponent-history variables as a secondary channel, and
modal-frequency tracking variables only weakly. Since the same pattern appears
for both Red and Black players, the behavioural ranking is unlikely to be a
role-specific artifact.

\paragraph{}The SER results show that these predictive gains are strategically meaningful. Lower SER means greater exploitability: a best-responding opponent makes fewer strategic errors when using the model's predictions. For both roles, SER falls steadily from the constant model through ME1 to ME2$^\ast$/ME3$^\ast$, approaching the LSTM benchmark (Table~\ref{tab:main-results}).

\section{Conclusion}\label{sec:discussion}

This paper asks whether the predictive power demonstrated by black-box sequence
models in repeated O'Neill play can be recovered by interpretable models. The
answer is largely yes. ME3$^\ast$ recovers about $89\%$ of the LSTM benchmark's
KL improvement for Red players and $98\%$ for Black players. However, whether the improved models actually decoded what was captured inside the black-box, or if they captured different regularities of the data and happened to perform similarly requires further testing.

\paragraph{}Regarding human's behavior, the main behavioural signal is repeat-and-avoid behaviour. Standard EWA explains
only a limited share of predictability, while adding own action-history terms
produces the largest improvement. Payoff-relevant opponent histories add further
predictive power, but modal-frequency tracking contributes little out of sample.
Thus, ME3 is valuable mainly as a diagnostic test: by mostly rejecting
best-response-to-mode behaviour as the missing mechanism, it strengthens the
repeat-and-avoid interpretation. The LASSO results also corroborate this ranking:
own repeat-and-avoid features are selected most consistently, opponent-history
features appear secondarily, and frequency-tracking features are weak.

\paragraph{} Overall, the results suggest that human randomisation fails in a structured
way, and what the interpretable models
managed to recover is not deep strategic sophistication, but rather a simpler vulnerability:
in covering their traces, players leave their own.

\begin{credits}
\subsubsection{\ackname} We are grateful to the authors
of~\cite{hirasawa2025} for generously sharing the experimental data, which
were collected as part of Professor Michihiro's game-theory course. 
\subsubsection{\discintname} The authors have no competing interests to
declare that are relevant to the content of this article.
\end{credits}

\FloatBarrier

\begingroup
\footnotesize
\setlength{\itemsep}{0pt}
\setlength{\parskip}{0pt}

\endgroup


\begin{thebibliography}{99}

\bibitem[Hirasawa et~al.(2025)]{hirasawa2025}
Hirasawa, T., Kandori, M., Matsushita, A.:
Using Big Data and Machine Learning to Uncover How Players Choose Mixed
Strategies. Working paper (2025)

\bibitem[O'Neill(1987)]{oneill1987}
O'Neill, B.:
Nonmetric Test of the Minimax Theory of Two-Person Zerosum Games.
Proceedings of the National Academy of Sciences
\textbf{84}(7), 2106--2109 (1987)

\bibitem[Brown and Rosenthal(1990)]{brown1990}
Brown, J.N., Rosenthal, R.W.:
Testing the Minimax Hypothesis: A Re-examination of O'Neill's Game Experiment.
Econometrica \textbf{58}(5), 1065--1081 (1990)

\bibitem[Rapoport and Boebel(1992)]{rapoport1992}
Rapoport, A., Boebel, R.B.:
Mixed Strategies in Strictly Competitive Games: A Further Test of the Minimax
Hypothesis.
Games and Economic Behavior \textbf{4}(2), 261--283 (1992)

\bibitem[Brown(1951)]{brown1949}
Brown, G.W.:
Iterative Solution of Games by Fictitious Play.
In: Koopmans, T.C. (ed.) \emph{Activity Analysis of Production and Allocation},
pp. 374--376. Wiley, New York (1951)

\bibitem[Roth and Erev(1995)]{rotherev1995}
Roth, A.E., Erev, I.:
Learning in Extensive-Form Games: Experimental Data and Simple Dynamic Models
in the Intermediate Term.
Games and Economic Behavior \textbf{8}(1), 164--212 (1995)

\bibitem[Camerer and Ho(1999)]{camerer1999}
Camerer, C.F., Ho, T.-H.:
Experience-Weighted Attraction Learning in Normal Form Games.
Econometrica \textbf{67}(4), 827--874 (1999)

\bibitem[Mullainathan and Spiess(2017)]{mullainathan2017}
Mullainathan, S., Spiess, J.:
Machine Learning: An Applied Econometric Approach.
Journal of Economic Perspectives \textbf{31}(2), 87--106 (2017)

\bibitem[Athey and Imbens(2019)]{athey2019}
Athey, S., Imbens, G.W.:
Machine Learning Methods That Economists Should Know About.
Annual Review of Economics \textbf{11}, 685--725 (2019)

\bibitem[Rudin(2019)]{rudin2019}
Rudin, C.:
Stop Explaining Black Box Machine Learning Models for High Stakes Decisions
and Use Interpretable Models Instead.
Nature Machine Intelligence \textbf{1}(5), 206--215 (2019)

\bibitem[Tibshirani(1996)]{tibshirani1996}
Tibshirani, R.:
Regression Shrinkage and Selection via the Lasso.
Journal of the Royal Statistical Society: Series B \textbf{58}(1), 267--288
(1996)

\bibitem[Hochreiter and Schmidhuber(1997)]{hochreiter1997}
Hochreiter, S., Schmidhuber, J.:
Long Short-Term Memory.
Neural Computation \textbf{9}(8), 1735--1780 (1997)

\end{thebibliography}
\end{document}